\documentclass[a4paper, 10pt, conference]{ieeeconf}      

\IEEEoverridecommandlockouts                              

\usepackage{graphics} 
\usepackage{graphicx}
\usepackage{epsfig} 
\usepackage{amsmath} 
\usepackage{amssymb}  
\usepackage{multirow}
\usepackage{subfigure}
\usepackage{mathptmx} 
\usepackage{amsmath} 
\usepackage{amssymb}  
\usepackage{breqn}
\usepackage{graphicx} 
\usepackage{amsmath,amssymb}
\usepackage{lipsum}
\usepackage{mathtools}
\usepackage{cuted}

\usepackage{xcolor}

\title{\LARGE \bf
Attention Augmented Convolutional Transformer for Tabular Time-series}

\author{Sharath M Shankaranarayana$^{1}$ and  Davor Runje$^{1,2}$ 
\thanks{$^{1}$ Airt Research, Zagreb, Croatia}
\thanks{$^{2}$ Algebra University College, Zagreb, Croatia}
}

\begin{document}
\maketitle
\thispagestyle{empty}
\pagestyle{empty}
\begin{abstract}
Time-series classification is one of the most frequently performed tasks in industrial data science, and one of the most widely used data representation in the industrial setting is tabular representation. In this work, we propose a novel scalable architecture for learning representations from tabular time-series data and subsequently performing downstream tasks such as time-series classification. The representation learning framework is end-to-end, akin to bidirectional encoder representations from transformers (BERT) in language modeling, however, we introduce novel masking technique suitable for pretraining of time-series data. Additionally, we also use one-dimensional convolutions augmented with transformers and explore their effectiveness, since the time-series datasets lend themselves naturally for one-dimensional convolutions. We also propose a novel timestamp embedding technique, which helps in handling both periodic cycles at different time granularity levels, and aperiodic trends present in the time-series data. Our proposed model is end-to-end and can handle both categorical and continuous valued inputs, and does not require any quantization or encoding of continuous features.     

\end{abstract}

\section{Introduction}

Industrial entities in domains such as finance, telecommunication, and healthcare usually log a large amount of data of their customers or patients. The data is typically in the form of events data, capturing interactions their users have with different entities. The events could be specific to the respective domains, for example, financial institutions log all financial transactions made by their customers, telecommunication companies log all the interactions of the individual customers, national healthcare systems keep logs of all visits and diagnoses patients had in different healthcare institutions, etc.
These kinds of data mostly employ tabular representation. A multivariate time-series data represented in tabular form is often referred to as \textit{dynamic} tabular data \cite{padhi2021tabular} or tabular time-series. These datasets are rich for performing knowledge discovery, doing analytics, and also building predictive models using machine learning (ML) techniques. Even though there exists a large amount of data, the data often remains unexplored due to various inherent difficulties (such as unstructuredness, noise, sparse and missing values). Because of the complexities, usually only a subset of the data is employed for building ML models. Thus, a large amount of potentially rich and relevant data is unused for ML modeling. Moreover, machine  learning  using  such  data is frequently performed by first extracting hand-crafted or engineered features and later building task-specific machine learning models. This engineering of the features is known to be one of the biggest challenges in industrial machine learning, since it requires domain knowledge and time-consuming experimentation for every different use case. Additionally, it has limited scalability because models cannot be automatically created and maintained. Another important issue is that most machine learning models require data to be in the form of fixed-length vectors or sequences of vectors. This is not straightforward and is difficult to obtain for such large time-series datasets. To this end, we propose a framework for learning vector representations from large time-series data. Our framework can later employ these learned representations on downstream tasks such as time-series classification.

 Learning numerical vector representations (embeddings), or representation learning, is one the major areas of research, specifically in the natural language processing (NLP) domain. In the seminal work called word2vec \cite{mikolov2013efficient}, vector  representations  of  words are learnt from  huge quantities of  text.  In word2vec, each  word  is  mapped  to  a $d$-dimensional  vector  such  that  semantically  similar  words have geometrically closer vectors. This is achieved by predicting  either  the  context  words  appearing  in  a  window around a given target word (skip-gram model), or the target word given the context (called continuous bag of words or CBOW model). In this model, it is assumed that the words appearing frequently in similar contexts share statistical properties and thus this model leverages word co-occurrence statistics. 
 
NLP has since then seen massive improvements in terms of results by building upon embeddings-related works. For example, the work \cite{howard2018universal} proposed a transfer learning scheme for text classification and thus heralded a new era by making the transfer learning in NLP on par with the transfer learning tasks in computer vision. Another recent work \cite{peters2018deep} proposed contextualized word representations, as opposed to static word representations.

The current state-of-the-art techniques in NLP learn vector representations of words using transformers and attention mechanism \cite{vaswani2017attention} on large datasets with the task of reconstructing an input text with some of the words in it randomly masked \cite{devlin2018bert}. 

\begin{figure*}[tp!]
  \includegraphics[width=\textwidth]{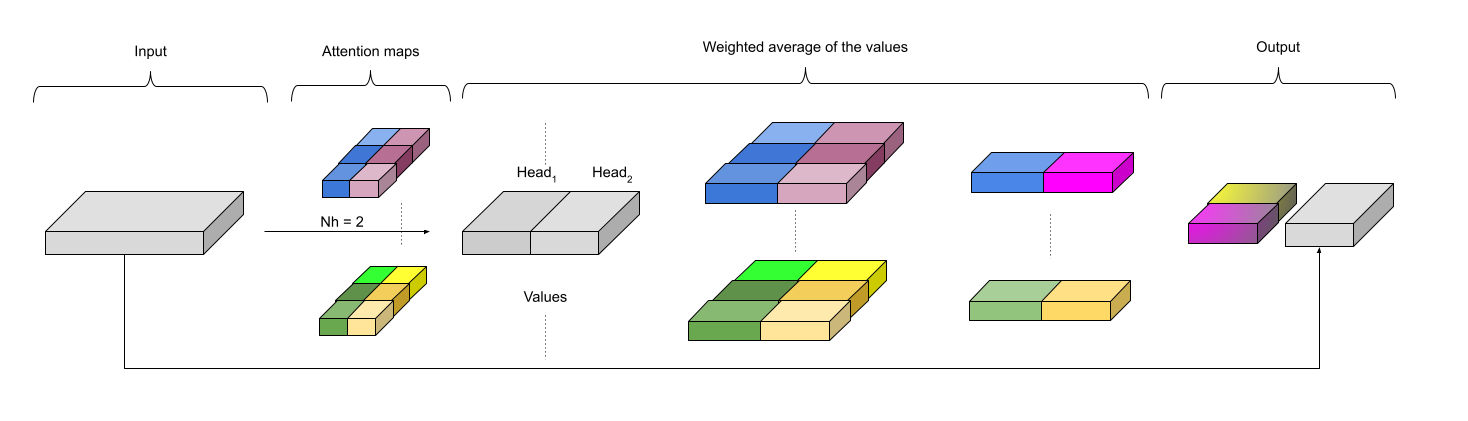}
    \caption{Attention-augmented convolution: For each temporal location, $N_h$ attention maps over the image are computed from
queries and keys. These attention maps are used to compute $N_h$ weighted averages of the values $V$. The results are then concatenated,
reshaped to match the original input’s dimensions and mixed with a pointwise convolution. Multi-head attention is applied in
parallel to a standard convolution operation and the outputs are concatenated.}
\label{AAconv}
\end{figure*}

 Upon seeing the immense gains of employing models with the attention mechanism and transformers in NLP, there have been a few works employing them for tabular data. In the work \cite{arik2019tabnet}, the authors propose a network called TabNet, which is a transformer based model for self-supervised learning for tabular data. They report significant gains in prediction accuracy when the labelled data is sparse, however, the work does not learn any sort of embeddings inherently and translates the use of transformer architecture from NLP domain to tabular domain. Another similar very recent work \cite{huang2020tabtransformer} proposes similar use of transformers for the tabular data. In this work, although the authors do intend to learn embeddings, it is only for encoding categorical features present in the tabular data. Whereas in our work, we learn the embeddings for specific agents by contextualizing the interactions between the various agents. Thus, the embeddings we obtain are very informative, unlike broad encoding of categorical values. Most recently, the authors in \cite{dang2021ts} proposed the use of BERT architecture modified for time-series anomaly detection task. 
 
 To the best of our knowledge, the closest work to ours is TabBERT proposed in the work \cite{padhi2021tabular}, in which the authors employ hierarchical BERT \cite{zhang2019hibert,pappagari2019hierarchical} to first perform pretraining on time-series data and then later employ the embeddings on downstream tasks. Although similar, our proposed work differs in several aspects:
\begin{itemize}
    \item We do not employ hierarchical BERT but single encoder layer consisting of transformers, and make use of embeddings to transform raw inputs in order to be fed to BERT, making its computational cost significantly smaller.
    
    \item We propose a masking technique specifically suited for time-series data resulting in higher performance of the model.
    
    \item TabBERT requires input data to be already encoded as categorical values, whereas our proposed framework can handle both discrete and continuous inputs. 
\end{itemize}

Another relevant work to ours is \cite{zerveas2021transformer}, where the authors propose transformers for time-series representation learning, but the pretraining loss function employed is mean squared error loss for both continuous and discrete features, whereas in our method we apply classification loss for discrete type features and regression loss for floating features. To the best of our knowledge, ours is the first work that employs attention augmented convolutions \cite{bello2019attention} (as a part of BERT layer) for time-series data and also the first work that proposes timestamp embedding block.

In summary, the main contributions of our paper are as follows:

\begin{itemize}
   \item We propose a novel BERT framework employing attention augmented convolutions for time-series tabular data TabAConvBERT.
   \item We propose a masking scheme that is better suitable for time-series tabular data.
  \item We propose a novel timestamp embedding block along with positional encoding. This timestamp embedding block helps in handling of both periodic cycles at different time granularity levels and aperiodic trends.
  \item The proposed framework can handle both discrete and continuous input types.
\end{itemize}

\vspace{0.7em}

\begin{figure*}[tp]
  \includegraphics[width=\textwidth]{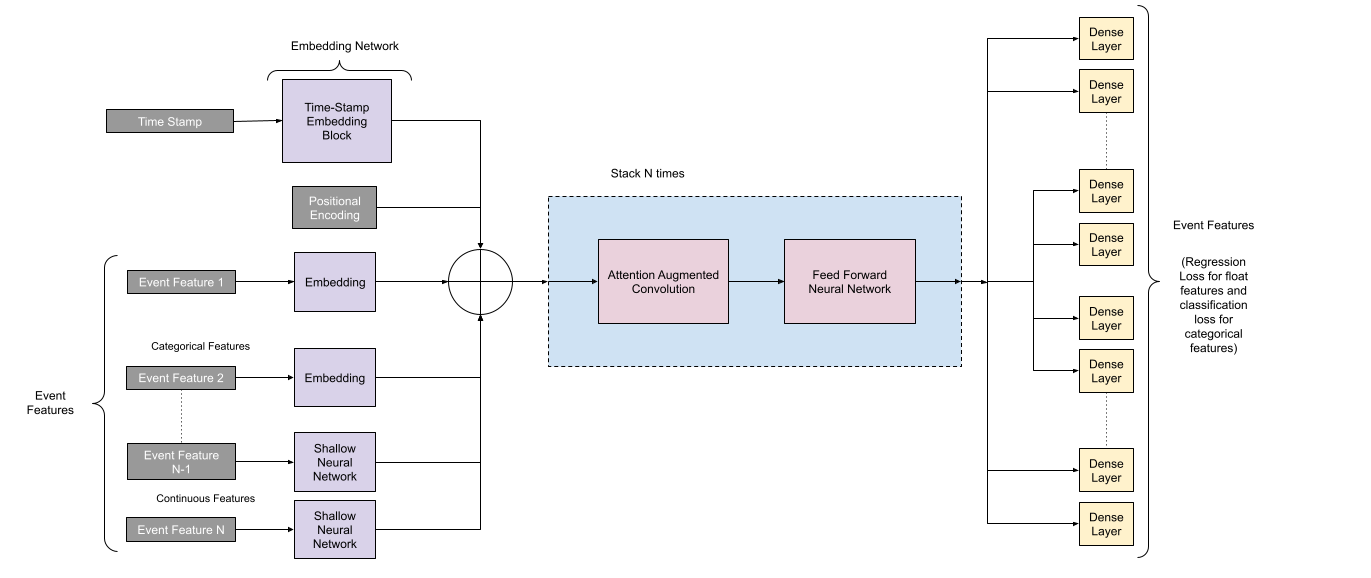}
    \caption{Our proposed architecture of TabAConvBERT}
\label{Architecture}
\end{figure*}
\section{Methodology}

\subsection{Attention Augmented Convolution}
One-dimensional (1-D) convolutions have long been employed in time-series tasks and have shown reasonably good results \cite{bai2018empirical}. In computer vision, employing attention mechanisms along with convolutional neural networks have given significant improvements in various vision tasks \cite{hu2018squeeze, hu2018gather}, \cite{park2018bam, woo2018cbam}. Recently, the authors of the work \cite{bello2019attention} proposed \textit{attention augmented convolutional networks} that can jointly attend to both spatial and feature subspaces, in contrast to previous convolutional attention mechanisms, which either perform only channel-wise reweighing \cite{hu2018squeeze, hu2018gather} or perform reweighing of  both channels and spatial positions independently\cite{park2018bam, woo2018cbam}. In this work, we propose a 1-D convolution version of attention augmented convolutions \cite{bello2019attention}. Similar to \cite{bello2019attention}, the proposed attention augmented convolution 
\begin{enumerate}
    \item is equivariant to translation, and
    
    \item can readily operate on inputs of different temporal dimensions. 
\end{enumerate}

As shown in \cite{bello2019attention}, if we consider an original convolution operator with kernel size $k$, $F_{in}$ input filters, and $F_{out}$ output filters, the corresponding attention augmented convolution can simply be written as:

\begin{dmath}\label{eq4}
 AAConv(X) = Concat[Conv(X), MHA(X)] 
\end{dmath}     
where $Conv$ denotes convolution and $MHA$ denotes multi-head attention.

\subsection{Architecture}

\begin{figure}[tp!]
\begin{minipage}{1.0\linewidth}
  \centering
  \centerline{\includegraphics[width=\textwidth]{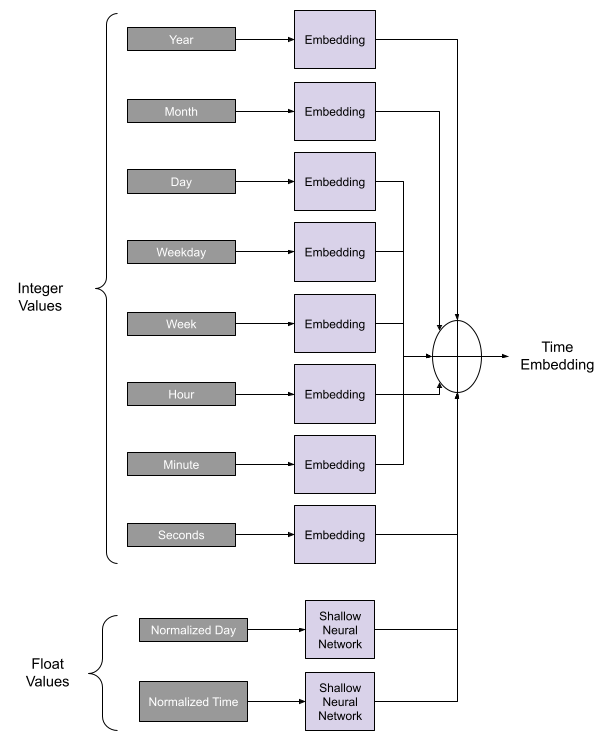}}
    \caption{Our proposed Timestamp Embedding Block}
    \label{TimeStamp}

\end{minipage}
\end{figure}

The proposed architecture is shown in Fig. \ref{Architecture}. The architecture consists of embedding network, which can encode both categorical inputs (by using simple embedding neural network) and continuous inputs (by using simple shallow neural network). The architecture also consists of a special timestamp embedding block, which is described in more detail in Fig. \ref{TimeStamp}. We first break the original raw timestamp into multiple components such as year, month, day, weekday, week, hour, minute and seconds. These broken up components are discrete and have a finite set of values. For example, month feature has $12$ values, week feature has $52$ values,  weekday feature has $7$ values, and so on. Each of these broken up discrete features are passed through an embedding layer. The outputs of each of the embedding layers are then summed-up. Additionally, we also create normalized timestamp features based on dates and time. These float values are then passed through a shallow neural network with “activity regularization” to obtain the vector representations having the same output dimensions of embedding representation. The resulting outputs from these shallow neural networks are also added to the summed-up embedding vector to obtain the final time embedding. The obtained time embedding is then added with input features' embedding and also positional encoding. The resulting summed value is then passed to attention augmented convolution layer and feed forward neural network layer. This combination of attention augmented convolution layer and feed forward neural network layer can be stacked $N$ times to obtain the encoder layer of BERT.  Finally, the dense layers in the output part of the network are added to aid pretraining or downstream tasks.

\subsection{Masking}
\begin{figure}[tp]
\begin{minipage}{1.0\linewidth}
  \centering
  \centerline{\includegraphics[width=\textwidth]{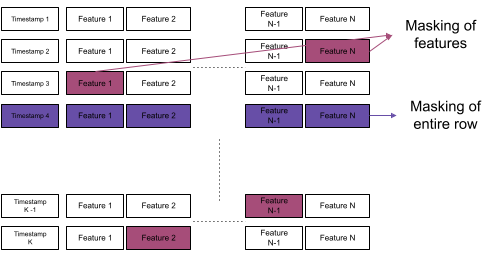}}
    \caption{Our proposed Masking methodology}
    \label{Masking}

\end{minipage}
\end{figure}

For learning representations from the time-series tabular data, we propose a procedure called masked data modeling (MDM), akin to masked language modeling (MLM) employed in NLP. One of the foremost steps in pretraining, as the name suggests, is masking. In MLM, masking is straightforward, since the languages contain only sequence of words. However, in multivariate time-series data, we propose two kinds of masking. For the first kind of masking, we mask out certain percentage of the features at random from the tabular time-series data. This masking is done independently for all the features and is similar to the one performed in \cite{arik2019tabnet}. For the second kind of masking, we randomly mask out certain percentage of entire rows of features for the time-series inputs. Almost all the previous works perform only first kind of masking for the tabular data. But only masking out individual features may not be very effective since often the features are slightly correlated and hence it becomes easier for MDM to predict missing features. The masking of entire row is similar to masking a word in MLM since, in tabular data, the analogue of a "word" is an entire row of features. The figure Fig. \ref{Masking} gives a clear picture of two types of masking employed in our work. 

Additionally, different from other previous works, our framework has the ability to handle continuous type of inputs as is, without resorting to binning. Although for masking the categorical data we can simply have a specific integer token similar to MASK token employed in MLM, for continuous type of inputs, we mask the data by replacing the original values with the mean value of the particular continuous feature. 

\section{Experiments and Results}

For our experiments, we employ the dataset provided in the work \cite{padhi2021tabular}. The dataset has 24 million transactions from 20,000
users. Each transaction (row) has 12 fields (columns) consisting of both continuous and discrete nominal attributes, such
as merchant name, merchant address, transaction amount, etc. For easier comparison, we employ the similar procedures for sampling as performed in  \cite{padhi2021tabular} by creating samples as sliding windows of 10 transactions, with a stride of 5. The strategies employed in masking are different since we propose custom masking methodology, however, we perform an experiment by quantizing and creating a vocabulary same as \cite{padhi2021tabular}. For masking, we perform $30\%$ masking of sample's fields and $15\%$ masking of sample's entire rows. While performing pretraining, we omit the label column to prevent biasing. In our proposed TabAConvBERT architecture, we only employ a single AAConvBERT layer to keep the number of parameters low.

For the downstream task of fraud classification, we again perform similar procedures by creating samples by combining 10 contiguous rows (with a stride of 10) in a time-dependent manner for each user, and thus obtain 2.4M samples, with 29,342 labeled as fraudulent. For evaluation, we use F1 binary score, on a test set consisting of 480K samples for better comparison. For the downstream task, we remove the final dense layers employed during the pretraining stage and a single dense layer for binary classification. For downstream classification task, after pretraining, we freeze all embedding layers and finetune the layers from attention augmented convolution layer. We perform three different kinds of experiments. First, we employ the same quantization and same masking employed in \cite{padhi2021tabular}, with the main difference being the BERT architecture having attention augmented convolutional layers. Next, we employ our proposed masking scheme but keep the same quantized data. This helps in gauging the improvements offered by our masking scheme. Lastly, we do not perform any encoding and keep all the continuous valued features as is, and also employ timestamp encoding as described in the previous section. Since we employ the same data partitions as described in \cite{padhi2021tabular}, we directly compare our results from their paper.

As seen from Table \ref{table1}, the proposed methods give significantly improved results on the downstream fraud classification task. The improvements can be seen by directly employing our TabAConvBERT architecture on the quantized or encoded data. The masking procedure also helps in providing a slight improvement in the F1 score. The best results can be seen from our TabAConvBERT with no data preprocessing, i.e., keeping the continuous and categorical values as is and employing timestamp embedding.

\section{Conclusion}
\begin{table}

  \centering
  \begin{tabular}{|c|c|}
    \hline
    \textbf{Method} & \textbf{Fraud F1 Score}\\
   
    \hline
     Baseline MLP \cite{padhi2021tabular} & 0.74\\ \hline
     Baseline LSTM \cite{padhi2021tabular} & 0.73 \\ \hline
     TabBERT MLP \cite{padhi2021tabular}  & 0.83\\ \hline
     TabBERT LSTM \cite{padhi2021tabular}  & 0.86  \\ \hline
     Proposed TabAConvBERT (Only Encoded Data) & 0.888\\ \hline
     Proposed TabAConvBERT (With custom masking) & 0.892\\ \hline
     Proposed TabAConvBERT & \textbf{0.896}\\ \hline
 \end{tabular}
    \vspace{1em}
 \caption{Comparison of various methods }
  \label{table1}

  \end{table}

In this work, we proposed a novel end-to-end BERT based architecture for time-series tasks. For the first time, we proposed the use of attention augmented convolutions for tabular time-series data and also proposed major modifications to the masking methodology for tabular time-series data. From our experiments, we showed that each of the individual modifications lead to improved results. Our method has a major advantage that it can be directly used with raw data, without resorting to techniques such as feature quantization or encoding. In future, we would like to rigorously evaluate on even larger scale industrial datasets.
\bibliographystyle{IEEEtran}
\bibliography{iscmi2021}

\end{document}